\documentclass[aps,pra,showpacs,preprintnumbers,amsmath,amssymb]{article}
\usepackage{arxiv}
\usepackage{mathptmx}
\usepackage{cite}  
\usepackage[pdftex]{graphicx} 
\usepackage{url}

\usepackage{times}
\normalfont
\usepackage[T1]{fontenc}

\def\hb{\hbox to 10.7 cm{}}

\title{
Loss aversion fosters coordination among independent reinforcement learners
}

\author{	
	Marco Jerome Gasparrini\\
	University Pompeu Fabra\\
	Barcelona, Spain	 
	\And 
	Mart\'i S\'anchez-Fibla\thanks{Corresponding Author: Technology Department, Universitat Pompeu Fabra, Carrer de Roc Boronat 138, 08018 Barcelona, Spain. E-mail:
		marti.sanchez@upf.edu.}\\
	University Pompeu Fabra\\
	Barcelona, Spain	 
}

\renewcommand{\headeright}{arXiv ver.}
\renewcommand{\undertitle}{arXiv version\thanks{Version of the extended abstract presented in the International Conference of the Catalan Association for Artificial Intelligence, CCIA 2018 and appeared in 
		Artificial Intelligence Research and Development: Proceedings of the International Conference of the Catalan Association for Artificial Intelligence, Volume 308, pags 307-311, IOS Press, 2018.}}

\begin{document}
	\maketitle

\begin{abstract}
We study what are the factors that can accelerate the emergence of collaborative behaviours among independent selfish learning agents. We depart from the "Battle of the Exes" (BoE), a spatial repeated game from which human behavioral data has been obtained (by Hawkings and Goldstone, 2016) that we find interesting because it considers two cases: a classic game theory version, called ballistic, in which agents can only make one action/decision (equivalent to the Battle of the Sexes) and a spatial version, called dynamic, in which agents can change decision (a spatial continuous version). We model both versions of the game with independent reinforcement learning agents and we manipulate the reward function transforming it into an utility introducing "loss aversion": the reward that an agent obtains can be perceived as less valuable when compared to what the other got. We prove experimentally the introduction of loss aversion fosters cooperation by  accelerating its appearance, and by making it possible in some cases like in the dynamic condition. We suggest that this may be an important factor explaining the rapid converge of human behaviour towards collaboration reported in the experiment of Hawkings and Goldstone. 
\end{abstract}

 \keywords{Multiagents, Reinforcement Learning, Game Theory}
	
\section*{Introduction}

There is a growing interest in the multiagent Reinforcement Learning (RL) community for spatial versions of Game Theory (GT) scenarios that can be seen as the first steps towards embodied ecologically valid setups in which agents learn to make decisions in complex environments, including complex reward contingencies \cite{perolat2017multi} and in the presence of other competitive learning agents \cite{leibo2017multi,lerer2017maintaining} (to name a few).  Recently economic notions, like loss aversion, have been incorporated to Reinforcement Learning settings \cite{hughes2018inequity} something that we also do in this paper with the objective to fit behavioural human data of \cite{hawkins2016formation} as we also started doing in \cite{freire2018modeling}. 

We study cooperation behaviours on a spatial version of an Iterated (also called Repeated) Continuous game called "Battle of the Exes" (BoE) (introduced in \cite{hawkins2016formation}) which is interesting because it considers two scenarios: a classic GT version, called ballistic, in which agents can only make one action/decision (equivalent to the classic GT game Battle of the Sexes) and a spatial version, called dynamic, in which agents can change decision (see figure \ref{fig:setup} and section \ref{setup} for details).

We model both versions of the game with independent RL agents and we introduce an utility function based on the obtained reward modified by a "loss aversion" factor: the reward that an agent obtains is perceived as less valuable when it is lower than what the other got or is getting. We see experimentally the introduction of "loss aversion" fosters cooperation by accelerating its appearance (meaning that agents learn to be fair sooner) and sometimes making cooperation possible, as in the dynamic case.

We first describe the BoE task in next section \ref{setup}, we then present the main concepts of the reinforcement learning algorithm used, the loss aversion based utility function and fairness in section \ref{methods} and we end with the results section \ref{results}.


\section{Setup}
\label{setup}

\begin{figure}[!t]
	\begin{small}
		\begin{center}
			\includegraphics[width=12cm]{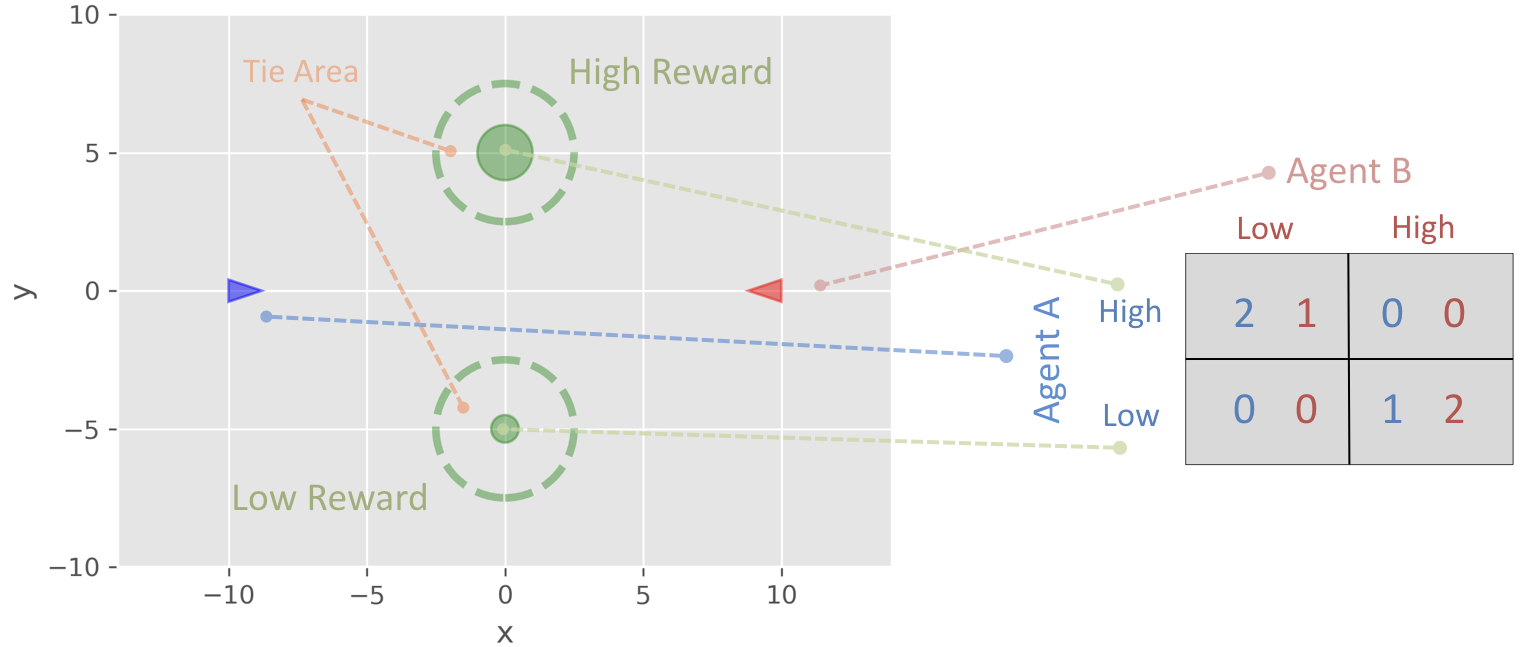} 
			\caption{Battle of The Exes (BoE) environment and payoff matrix. Left: The environment is depicted. Positions $p_a = \langle -10,0 \rangle$ and $p_b = \langle 10,0 \rangle$ are the starting positions of the blue agent (player $a$) and red agent (player $b$) respectively. The high and low reward spots are placed in $p_{high} = \langle 0,5 \rangle$ and $p_{low} = \langle 0,-5 \rangle$ positions respectively.
			Right: Payoff matrix associated to ballistic version of the BoE game that also applies, somehow, to the dynamic (in this case a tie is defined dependent on the tie area).}
			\label{fig:setup}
		\end{center}
	\end{small}
\end{figure}

The Battle of the Exes (BoE)\cite{hawkins2016formation} is a Continuous Iterated Game in which two agents with constant speed can move freely changing its direction in an environment where two different valued rewards (high and low reward) are equidistant to the agents at the start of a trial. When an agent reaches a reward spot it is rewarded with its value (high or low) if the other agent is not inside what we call the "tie area", in which case both agents get nothing (see figure \ref{fig:setup} and references \cite{hawkins2016formation,freire2018modeling} for a detailed description of the game ). The game is played repeatedly. BoE has also a non continuous version called ballistic (a classical iterated game version): 
\textbf{(a)} Ballistic (Iterated Game): decisions are taken beforehand (before a trial starts) and outcomes of the trial are executed without possibility of changing decision, that is, ballistically. This version is equivalent to the classic games: Battle of the Sexes, Bach or Stravinsky and Hero.
\textbf{(b)} Dynamic (Continuous Iterated Game): players can change decision (that is approach the high or low reward) with no impact until the end of the trial when a reward is reached (that is when one of the players reaches a reward spot). 

\section{Methods}
\label{methods}

We model both the ballistic and dynamic versions of the BoE task with independent Reinforcement Learning (RL) agents having access only to the precedent episode outcome (high reward, low reward, tie) and to the $y$ position of all agents in the dynamic case. In this sense, we are modelling also the dynamic changes of decisions something that we did not address in a precedent paper \cite{freire2018modeling}. 
For the RL algorithm we use Q learning, which is an off-policy model-free algorithm with an epsilon greedy policy. Q-learning maintains a function (in our case there will be two functions, one for each agent), $Q(s,a)$, that given a state and an action, returns the future discounted reward that one can obtain from applying that action in that state. The Q function is updated using the following equation derived from the bellman equations:

$$
Q(s,a) \leftarrow Q(s,a) + \mu(r^i+\gamma \max_{a'} Q(s',a')-Q(s,a))
$$

where $Q^i$ is the Q function of agent $i$, $r_i$ its immediate reward, $s$ the current state, $a$ an action, $\mu$ the learning rate, and $\gamma$ the discount factor. In addition we need to specify the epsilon decay which corresponds to the decay in which the $\epsilon$-greedy policy will decrease the probability of executing a random action, that accounts for exploratory behaviours.

Each game (or also called session) between two players, has three possible convergence outcomes: dominant (the same agent gets the high reward and the other the low one), being dominated and turn taking (agents take turns to get the high reward). These outcomes correspond the three Nash equilibria of the game in which neither of the players has interest in changing strategy given the current one.

One factor that may be fostering turn taking in humans and lowering the number of episodes needed to achieve it, is the fact that reward is perceived differently in social contexts in which the obtained reward is compared among rewards obtained by others. Receiving the same amount of reward than a previous trial may be perceived differently if this reward is lower than what the other got (loss aversion), or simply it may happen that an inequity aversion principle is present because of social and cultural constructs. For this purpose we introduced a manipulation of the reward function in the form of an utility in a similar way than \cite{hughes2018inequity}. Agents perceive reward in the form of utility:

$$U_i(r_i,r_j,r^{ref}_i,r^{ref}_j)=r_i - \alpha \max(0,r^{ref}_j-r^{ref}_i)-\beta \max(0,r^{ref}_i-r^{ref}_j)$$

where $r_i,r_j$ are the rewards received by agent $i$ and $j$ respectively, $r^{ref}_i,r^{ref}_j$ are the reward references of each agent consisting of the sum of the last two rewards, $\alpha,\beta$ are the disadvantageous/advantageous inequity parameters respectively (see \cite{hughes2018inequity}). The fact that $r^{ref}_i$ considers the rewards of the last two episodes of agent $i$ differs from \cite{hughes2018inequity} and it was important for the obtained results.

The utility function always devalues the reward (by subtracting a scaled difference of what the agents got) or it leaves it untouched. We made all inequity parameters $\alpha$,$\beta$ not dependent on the agent and also we made $\alpha$ greater than $\beta$ to account for loss aversion. A pure loss aversion scenario is when $\beta$ equals 0.
In this case the reward function is not modified for agent $i$ when agent $i$ is earning more than $j$ being not sensitive to advantageous inequity. We show in the results section \ref{results} that loss aversion is sufficient to foster coordination. Note that setting $\alpha$ and $\beta$ to zero, leaves the reward unaltered.

\begin{figure}[!t]
	\begin{small}
		\begin{center}
			\includegraphics[width=10.5cm]{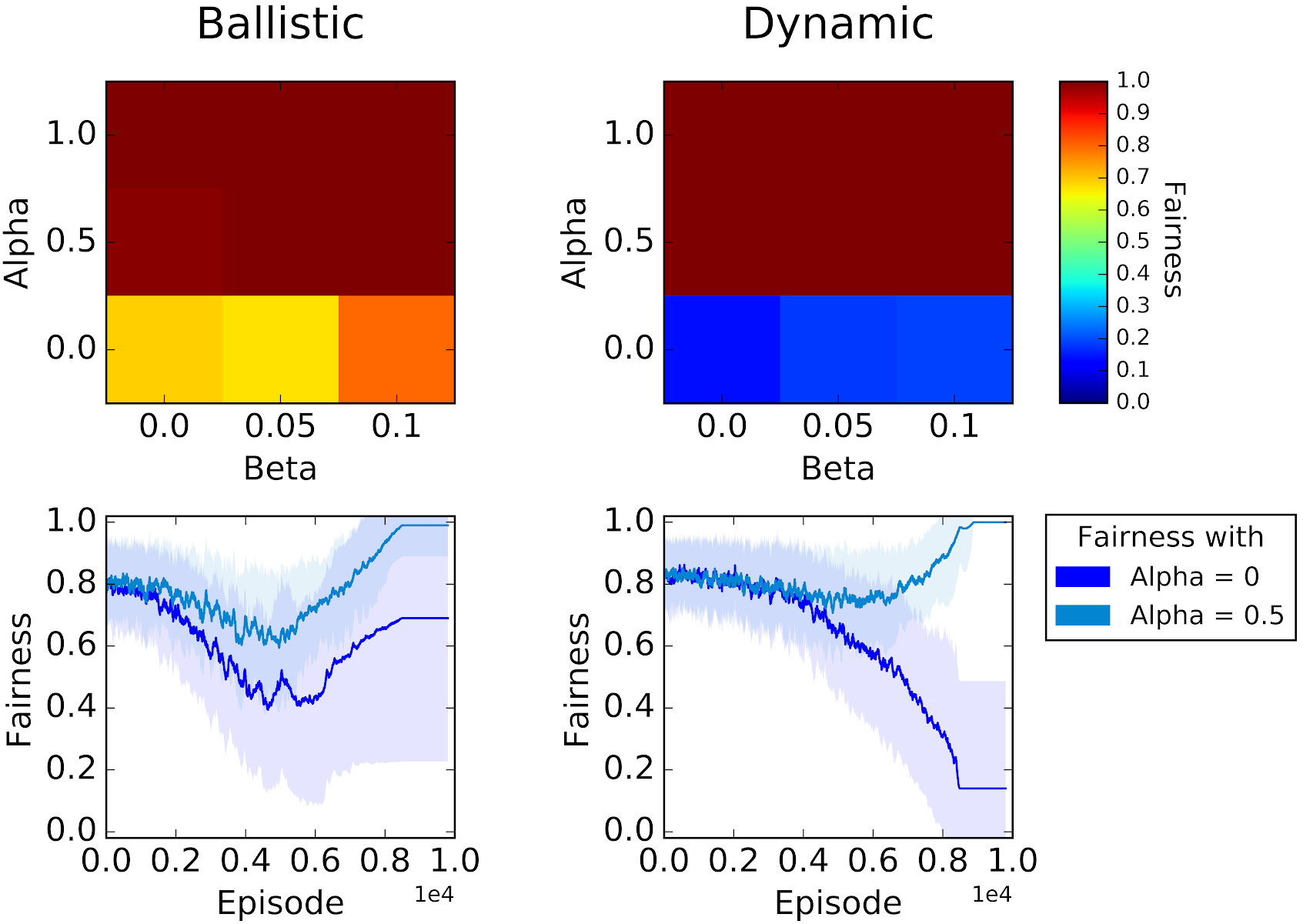} 
			\caption{Fairness in the $\alpha$,$\beta$ parameter space. Top Row: Fairness for values of $\alpha$,$\beta$ parameter space in the ballistic (left) and dynamic (right) conditions. Bottom Row: Fairness mean and standard deviation comparison across episodes for $\alpha=0$ and $\alpha=0.5$ of ballistic (left) and dynamic (right) conditions ($\beta$ is always 0 in these cases). Simulations parameters of all plots: 10000 episodes, 100 dyads, learning rate $\mu=0.3$, discount factor $\gamma=0.9$, linear epsilon decay that starts at probability 1 of making random actions and ends at episode 8500.} 
			\label{fig:results}
		\end{center}
	\end{small}
\end{figure}

We studied turn taking behaviours in \cite{moulin2015autonomous} in the context of marmoset producing vocalizations to maximize individual presence assessment in occluded environments. Here turn taking is characterized by computing fairness (as done in \cite{hawkins2016formation}). Fairness accounts for how well the total obtained reward is distributed among players (the minimum divided by maximum reward obtained by the players) and ranges from 0 to 1 (0 fairness means that all reward was unevenly awarded to one player and fairness 1 meaning that reward was evenly distributed, that is, each player got half of the total reward). We explain in the following how we compute fairness. Let $\{h_i^t | t=1...T\}$ be the sequence of 0,1 values indicating whether the agent got the maximum reward on that trial. Let $H_i^T = \sum_{t=1}^T{h_t^i}$ be the total times agent $i$ obtained the highest reward. Then the fairness would be computed as follows: $F^T = \frac{min\{H_a^T, H_b^T\}}{max\{H_a^T, H_b^T\}}$ and if nobody got the high reward during $T$ episodes the fairness is  assigned to $1$ to avoid division by $0$.

\section{Results}
\label{results}

First we study independent Q-Learning agents with no reward manipulation ($\alpha=0$ and $\beta=0$), but in this case we don't have conclusive results for the study of turn taking outcomes in the complete parameter space of $\gamma$ (discount factor) and number of episodes.
We are more interested in the effects of introducing loss aversion, thus we take a look at the $\alpha$,$\beta$ parameter space. In summary loss aversion clearly accelerates the appearance of turn taking: see the transition to 1 of the fairness when $\alpha$ changes from 0 to 0.5 in both ballistic and dynamic conditions (see figure \ref{fig:results}).

In the case of no reward manipulation ($\alpha=0$ and $\beta=0$), we find\footnote{These results are in progress and will be published in the master thesis of Bernat Puig supervised by Vicen\c c Gomez and Mart\'i S\'anchez Fibla} that turn taking behaviours can always be achieved in the ballistic case with a high discount factor $\gamma > 0.995$ and low learning rates $\mu < 0.5$ and a sufficiently large number of episodes $> 100000$. This high number of episodes is in dis-accordance with the low number of episodes needed by human participants to achieve turn taking (around 25, see \cite{hawkins2016formation}). 

In the dynamic condition without loss aversion we observed a peculiar result of the learning: agents learn to delay the decision by infinitely switching actions. This made the episodes longer and longer obliging us to set a limit on the maximal number of time steps per episode. A dilemma appears: which reward do we assign to an exceeding time step episode? We decided to assign 0 reward in this case, but we are not sure of its consequences (in the case of loss aversion we don't observe this). We also tested with higher $\gamma$ of $0.99$ and turn taking completely disappeared.
This result is also contradicting human experiments where turn taking increases in the dynamic version \cite{hawkins2016formation}.

\subsection{Introducing loss aversion} 
\vspace{-0.2cm}
We then look at the $\alpha$,$\beta$ parameter space considering only loss aversed agents: that is when $\alpha > \beta$ (results should be symmetric). Results show that turn taking behaviours appear in less episodes (see curves in the bottom row of figure \ref{fig:results}) and are more usual, being the only possible outcome in many regions of the $\alpha,\beta$ parameter space. Situations in which the two agents tie forever are also possible but may be an artifact of insufficient number of episodes. 

These results are promising because a minimal transformation of the reward function can accelerate the appearance of turn taking outcomes (in terms of fairness) and is a first direction to fit the human experimental results of \cite{hawkins2016formation}.

\vspace{.3cm}
\noindent \textbf{Acknowledgments}
\begin{small}
Research supported by INSOCO-DPI2016-80116-P. We thank Bernat Puig for sharing results of his master thesis. We also thank Cl\'ement Moulin-Frier, Vicen\c c Gomez, Ismael Tito Freire, Xerxes Arsiwalla and Marco Fongoni for helpful discussions.
\end{small}  

\vspace{-.3cm}

\bibliographystyle{plain}
\bibliography{LossAversion}

\end{document}